\documentclass[a4paper, 10pt, twocolumn]{article}

\usepackage{amsfonts,amssymb,amsmath,amsthm}
\usepackage{subfigure}
\usepackage{graphicx}
\usepackage[footnotesize]{caption}

\usepackage[top=1.5cm, left=1.5cm, right=1.5cm, bottom=1.5cm]{geometry}

\title{Lightweight U-Net for High-Resolution Breast Imaging}

\author{Mickael Tardy$^{1,2}$, Diana Mateus$^2$.\\
\footnotesize $^1$Ecole Centrale de Nantes, LS2N, UMR CNRS 6004, Nantes, France.\ $^2$Hera-MI, SAS, Nantes, France.
}
\date{\empty} 

\renewenvironment{abstract}{\bf\small {\em\ Abstract---}}{}

\begin{document}

\maketitle

\begin{abstract}
We study the fully convolutional neural networks in the context of malignancy detection for breast cancer screening. We work on a supervised segmentation task looking for an acceptable compromise between the precision of the network and the computational complexity. 
\end{abstract}

\section{Introduction}
\label{sec:introduction}

Breast cancer is one of the most prevalent cancers worldwide with 1/8th of the women population affected by the disease \cite{Siegel2019}. Regular breast cancer screening plays an important role in early breast cancer detection and mammography remains the most used around the world. Its main purpose is the discovery of the signs of cancer development (e.g. clusters of microcalcifications, spiked masses, distortions, etc.) leading to further examinations such as ultrasound, MRI, biopsy. Given the importance of breast cancer for public health, the area is one is most active in medical imaging the research \cite{Carneiro2017, Hamidinekoo2018a}.

The mammography analysis presents by its nature two main tasks: classification and detection. The difficulty of the tasks is emphasized by the complexity of the visualized features. That is, the image presents a projection of several layers of soft tissues. Moreover, the malignant findings sizes vary, with the smallest being below $0.5mm$ \cite{Mercado2014}, making the detection task even harder. 

Deep learning techniques are widely used for different medical-imaging-related tasks and mammography is amongst the areas of interest \cite{Carneiro2017, Hamidinekoo2018a}. However, the size and the complexity of the imaging makes the application of the state-of-the-art deep learning methods less straightforward, both in training and in test phases.

In the present work, we focused on the supervised segmentation task. We aim at maximizing the input resolution of the images to increase the detection capabilities while having explicit hardware limitations (i.e. one mass-market GPU).

\section{Related work}
\label{sec:intro}

Several light implementations of U-Net-type networks have been proposed \cite{Chen2018,Qi2019}. Sun et al.  proposed a U-Net for mammographies \cite{Sun2018} that involves downscaling image to (256x256), where it is difficult to detect certain suspicious regions (i.e. calcifications clusters). Oktay et al. \cite{Oktay2018} and Sun et al. \cite{Sun2018} propose structural modifications to a U-Net with attention layers, hence more complexity, while De Moor et al. propose to use a patch-wise trained U-Net on the full-sized mammographies \cite{DeMoor2018}. While it allows the processing of high-resolution images, the training process looses spatial and topological information of the full image. Pandey et al. introduced a network similar to ours, but the authors experiment on low-resolution imaging (256x256)  \cite{Pandey2018}. In most of the cases, the proposed lightweight architectures have been applied to solve tasks where the traditional U-Net is applicable and yields reasonable results. In contrast, we tempt to solve the task where regular U-Net does not fit.

\section{Methods}
\label{sec:format}

\begin{figure}[thb]
\centering
\includegraphics[width=6cm]{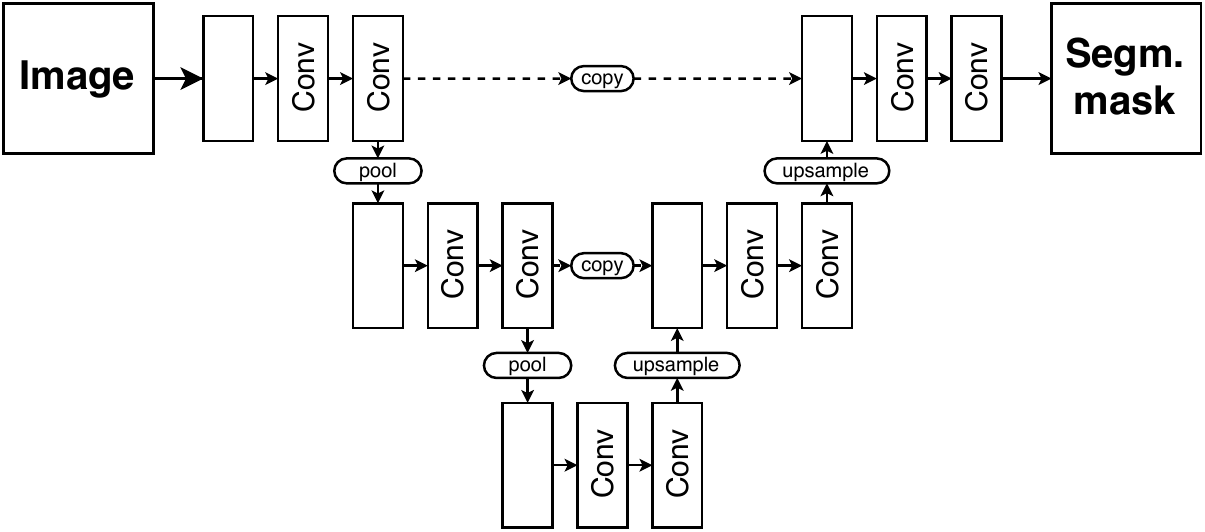}

\caption{Simplified illustration of the U-Net architecture having depth of 2 levels. For details, see \cite{Ronneberger2015}}
\label{fig:unet}
\end{figure}

Focusing on the segmentation task, we based our work on the U-Net, state-of-the-art deep learning architecture for segmentation \cite{Ronneberger2015} (see 2-levels-depth U-Net illustration on fig. \ref{fig:unet}), bringing several modifications. 

The initial U-Net processes images of 572x572 pixels and has 4-level-depth having  $\approx 10M$ parameters. The U-Net architecture being fully convolutional by design, it can process images of different sizes and in such cases, the limits are imposed by the hardware. 

As shown by \cite{DeMoor2018}, the training operations may be performed patch-wise on GPU hardware, and the testing is done image-wise (on CPU and RAM). While it allows running the training on images of smaller size, it looses spatial and topological information of the full image.

Unlike \cite{DeMoor2018} we aimed at training on full mammography images. To achieve our goal, we build a lightweight U-Net having several modifications compared to the initial U-Net. 

Having images of higher resolution, we increased the depth of the network from four to seven levels. That yielded a network with 600M parameters. To cope with such complexity, we decreased the number of the initial level convolution filters from 64, as in the original paper, to 16, which limited the number of parameters to  37M. To simplify the network even further, we replaced traditional convolutional layers with separable convolutions; that brought the number of parameters down to 4.6M. Finally, to mitigate the precision loss due to the limited number of parameters, we added short residual connections \cite{Chen2018} at each level of the U-Net, leading to the final 5.3M parameters network. 

We trained our network on full mammographies resized to 1536x1536. Such resolution yielded images of $\approx 0.15mm$ pixel spacing, which is acceptable with regards to the size of the malignant features \cite{Mercado2014}. Therefore, given the resolution of the input images, unlike most of the state-of-the-art approaches, our proposed network is trained to segment both, masses and calcifications. 

\section{Results}

We achieve $DICE = 0.58$, on the INbreast \cite{Moreira2012a} database, which is comparable to the state-of-the-art performances (0.62, \cite{Sun2018}), considering our higher scale and combination of masses and calcifications.

\begin{figure}[thb]
\centering
\includegraphics[width=8cm]{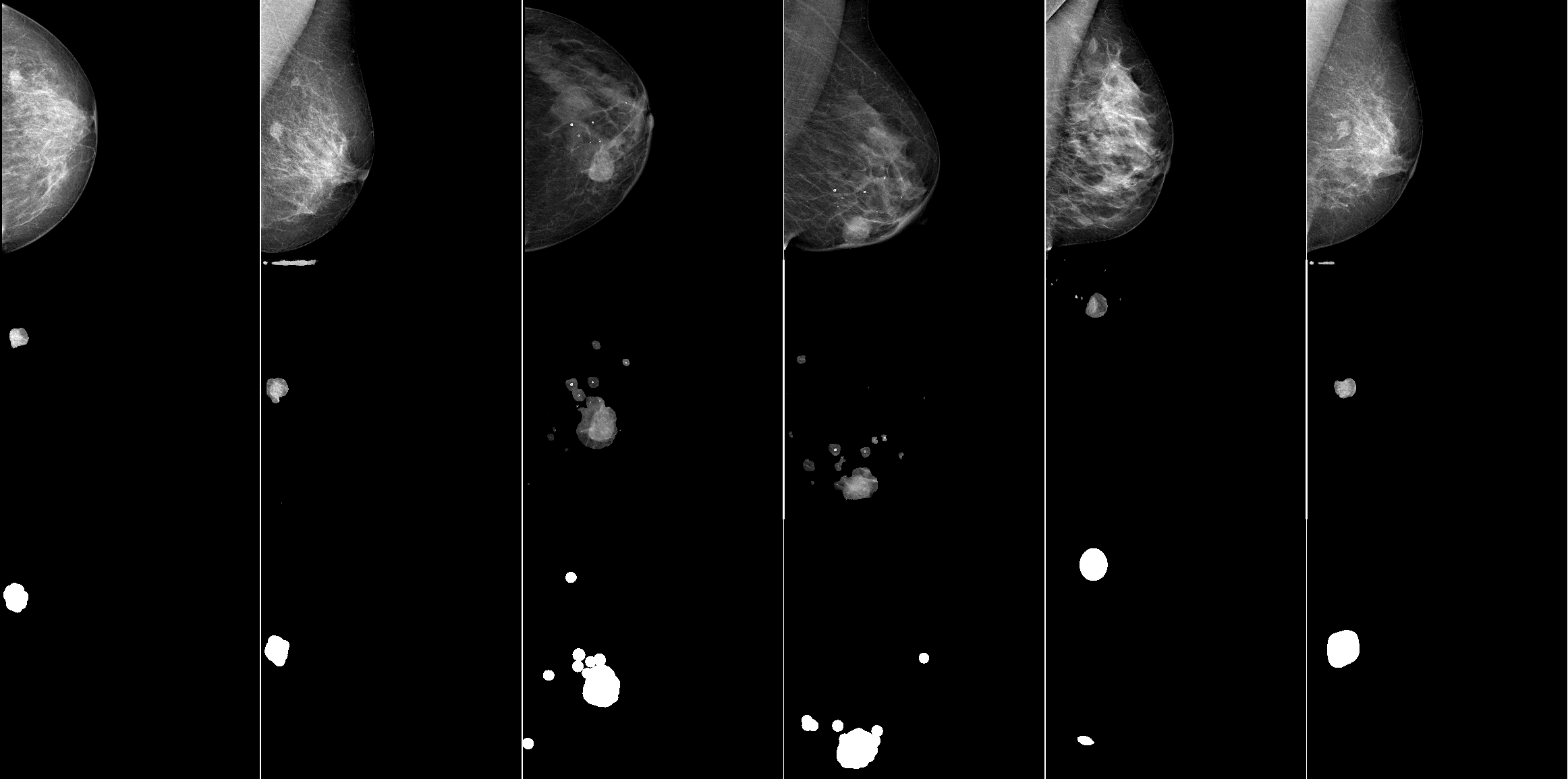}

\caption{Segmentation results of the proposed network. \textbf{First row}: input images; \textbf{Second row}: Thresholded output, \textbf{Third row}: ground truth (masses and mircocalcifications)}
\label{fig:res}
\end{figure}

\section{Discussion}
\label{sec:pagestyle}

In the present work, we focused on the mammography imaging segmentation task, looking for a compromise between the solution precision and complexity. We approached the problem with deep learning fully supervised method, based on state-of-the-art U-Net architecture.

To achieve our goal, we designed a network with a limited number of parameters so it can fit one mass-market GPU. Thanks to its lightness, our network can run fast in the test environment, which makes it production-ready.

The results obtained with our mammography segmentation deep learning network may be used in the screening process in two ways:
i) Guiding the radiologists during diagnosis and ii) Guiding the radiographers in decision for additional examinations before the images are shown to the radiologist.

Finally, our approach may be used for other types of imaging as well where both, high resolution and lower complexity are equally important. More precise architecture adjustments may be necessary for any particular task.

\bibliographystyle{abbrv}
\bibliography{Hera-MI-iTWIST2020.bib}

\end{document}